\documentclass[sigconf]{acmart}

\usepackage{colortbl}
\usepackage{booktabs}
\usepackage{amsmath}
\usepackage{graphicx}
\usepackage{amsfonts,bm}
\usepackage{multirow}
\usepackage{color}
\usepackage{caption}
\usepackage{mathtools}
\usepackage{enumitem}
\usepackage{makecell}

\setcopyright{none}
\renewcommand\footnotetextcopyrightpermission[1]{}
\begin{document}

\title[EvoBench: Evolving Diagnostic Benchmarks for Environmental Modeling]{EvoBench:
An Evolving, Diagnostic Benchmark Framework for Environmental Modeling %, with a Streamflow Case Study
}

\author{Yiming Sun$^1$, Yuan Zhang$^2$, Qi Cheng$^1$, Yingda Fan$^1$, Runlong Yu$^3$, Yiqun Xie$^4$, Xiaowei Jia$^1$}
\affiliation{%
  \institution{$^1$Rutgers University, New Brunswick\\ $^2$Apple Inc. \\$^3$University of Alabama\\ $^4$University of Maryland, College Park}
  \country{}
}
\renewcommand{\shortauthors}{Sun et al.}

\begin{abstract}
Machine learning models are now widely used to predict environmental variable dynamics in domains like hydrology, climate science,  agriculture, and ecology.  These models are
usually compared on fixed benchmark datasets using a single average accuracy
score. As models improve, their scores become very close when averaged over large regions and long time periods, so the benchmark can
no longer tell them apart, and a single number hides \emph{where} and
\emph{when} a model fails. Three failures are common across environmental
domains: models get worse as the climate shifts (non-stationarity), models fail
at unmonitored sites (spatial extrapolation), and models miss rare but important
regimes such as floods and droughts.

We present \textbf{EvoBench}, a benchmark framework that shows where a model fails, tracks failures over time, and evolves to focus on model capability gaps.
%is \emph{diagnostic} (it shows where a model fails) and \emph{evolving} (it keeps tracking failures over time). 
% EvoBench has five general parts: (i) a skill score measured against
% a re-runnable reference model, so results are comparable across different conditions; %easy and hardyears; 
% (ii) ``failure cells'' that break performance down by site attributes and
% by regime; (iii) a test of prediction at unmonitored sites; (iv) a loop that
% slides a time window forward and tracks which failures remain; and (v) a small,
% carefully scoped language-model agent, backed by a physical consistency check,
% that tells a true model weakness apart from bad data. 
We apply EvoBench to daily
streamflow prediction as a case study (\textbf{CAMELS-Evolve}, about 9{,}000 U.S.
basins, 1985--2024) and compare an operational physical model (NWM), a
conceptual reference (HBV), and several data-driven models. No model wins on every testing scenarios (e.g., in terms of %: %question: NWM has the best average
average accuracy, performance at low flows and unmonitored basins). %the LSTM is best at low flows, the EA-LSTM generalizes best to
%unmonitored basins, the Transformer is strong only at scale, and a
Also we observe that all data-driven
models under-predict floods. Additionally, we  show %the findings are actionable: a
that targeted fix based on the reported failures could mitigate weaknesses the benchmark identified in certain scenarios. %, while the same fix does
%not help a different weakness whose cause is different. 
EvoBench turns model
evaluation from a saturated ranking into a moving map of model limitations.
\end{abstract}




\maketitle

\section{Introduction}
\label{sec:intro}

Machine learning (ML) has been widely used in environmental modeling to 
predict streamflow, greenhouse-gas fluxes, air quality, soil moisture, and crop
yield, and they are often reported to exceed traditional process-based models in terms of both accuracy and computational ~\cite{kratzert2019, tran2025camelsh,sun2026x,chen2023physics,yu2025physics}. To compare these models,
researchers have built large benchmark datasets with fixed training and test
splits, and they usually report a single accuracy score such as RMSE, 
Nash--Sutcliffe Efficiency (NSE), and KGE.

This way of evaluating models has several issues. 
As models improve, their scores
become very close to one another. When many models reach nearly the same number,
the benchmark \emph{saturates}: the score can no longer tell good models from
bad ones. 
More importantly, despite a better overall score, ML models are known to struggle in out-of-distribution and data-sparse scenarios, which could be obscured by averaging performance over large regions and long time periods. For scientific use, knowing the failures matters more
than knowing the average. 
 %More importantly, a single average score hides \emph{where} and
%\emph{when} a model fails. 

Three types of failure are particularly common in machine learning applications across environmental domains. The first is \emph{non-stationarity}: a model that works
well on past data can get worse as the climate and land surface change, but a
random train/test split hides this because it mixes easy and hard years
together. The second is \emph{spatial extrapolation}: sensors are sparse and
unevenly placed, so a model that does well at monitored sites may fail at
unmonitored ones, yet most benchmarks only test monitored sites. The third is
\emph{regime-specific error}: because models are trained to minimize average
error, they tend to miss rare but important regimes such as floods and droughts,
which contribute little to an average score. A useful benchmark should therefore
be \emph{diagnostic} (it should show where a model fails) and \emph{evolving}
(it should keep tracking failures as models and the environmental conditions change). Since these problems are shared across many environmental domains~\cite{karpatne2018machine,yu2025survey}, the benchmark should be designed to be generalizable and adaptable across domains.
%Becausethese problems are shared across environmental domains, o should begeneral rather than tied to one task.

In this work, we introduce \textbf{EvoBench}, a diagnostic and evolving benchmark
framework for environmental modeling. We then apply it to daily streamflow
prediction as a detailed case study, which we call \textbf{CAMELS-Evolve}. We
build the case study on the CAMELSH dataset~\cite{tran2025camelsh}, which
provides daily weather forcing, observed streamflow, and catchment attributes
for about 9{,}000 river basins across the contiguous United States from 1985 to
2024. Hydrology is a good first case for three reasons: it shows all three
failure modes above, it has strong physical and operational baselines to compare
against, and new observations are being collected over time and can be used as ground truth. Unlike benchmarks in language modeling and computer vision,
environmental ecosystems keep producing new test data. Every new day of observed streamflow is an
example that no model could have seen during training. This makes it natural to
test models in a rolling, forward-in-time fashion, which better reflects real-world deployment and reduces the risk of benchmark overfitting.
Moreover, over the rolling evaluation, the benchmark is expected to  actively find the conditions under which models fail — certain basin types, flow regimes, or time periods — instead of reporting one overall performance score. 
Later rounds then concentrate on these open problems rather than rewarding cases that models have already solved. In this way the benchmark keeps steering attention toward the frontier of what current models cannot yet do.
%EvoBench therefore groups each round's errors into failure cells, checks that a failure reflects a true model weakness rather than bad data, and promotes the cells where models still struggle. Later rounds then concentrate on these open problems rather than rewarding cases that models have already solved. In this way the benchmark keeps steering attention toward the frontier of what current models cannot yet do.

EvoBench has several components, each designed to apply generally to diverse environmental-modeling
tasks. %, not just hydrology:

\begin{itemize}[leftmargin=1.2em]
\item \textbf{Comparable scores across time.} Some years are more challenging to
predict than others. To make scores comparable, we measure each model as a
\emph{skill} score relative to a reference model that can be re-run in any period. In
the case study, the reference is a simple calibrated hydrology model (HBV), which % that, unlike the operational National Water odel, 
can be run on any future input.

\item \textbf{Failure cells.} We break overall performance into small groups of test samples,
first by site attributes and then by within-record regime (e.g., low
versus high flow). This enables performance analysis on %turns a vague statement like ``the model is weak in
%extremes'' into 
a specific and trackable unit.

% \item \textbf{Spatial extrapolation.} We hold out whole groups of sites and
% test on them, which measures how well a model predicts at unmonitored locations.

\item \textbf{An evolving loop.} We move a time window forward in steps,
promoting the failure cells that remain unsolved and retiring the ones a model
has solved.

\item \textbf{An agentic guardrail.} A carefully scoped use of a language
model, backed by a model-independent physical check, decides whether a failure
is caused by a true model weakness or bad data.
\end{itemize}

In this case study, we use EvoBench to several physical models (NWM, HBV) and data-driven models (e.g., LSTM, EA-LSTM~\cite{kratzert2019towards}, Transformer, {TCN}, DLinear~\cite{zeng2023transformers},
TSMixer~\cite{chen2023tsmixer}, and PatchTST~\cite{nie2022time}). %In  five models: an operational physical model (NWM), the
%HBV reference, and three data-driven models (an LSTM, an EA-LSTM, and a small
%Transformer). 
We find that no single model is best at everything. The LSTM and GRU surpass NWM in average NSE and the LSTM is also best at low flows, the EA-LSTM generalizes best to unmonitored basins, general-purpose time-series architectures fall near or below a simple conceptual model, and all data-driven models badly under-predict floods, where NWM remains far ahead. 
We also perform two additional tests for these baseline models, by (1) iteratively improving the model using evaluation feedback from each round to identify and address its specific weaknesses, 
%2. try to fix the model with feedback in each round
%3. try model with memory over different evolving rounds
%and 
and (2)  incorporating memory across successive rounds, enabling it to retain insights from previous evaluations and refinements and leverage them to guide subsequent improvements.

Finally, we
show that the benchmark's findings are useful in practice: when it points to a
specific weakness (dry and snowy basins), one of two plausible targeted fixes
(over-sampling those basins) clearly improves both cells at no cost to overall
accuracy, while the other (re-weighting low flows in the loss) does not help. In short, EvoBench turns model
evaluation from a single, saturated ranking into a moving map of model
limitations.

\section{Related Work}
\label{sec:related}

\paragraph{Benchmarks for environmental modeling.}
Large-scale datasets and benchmarks now exist for many environmental problems, including
streamflow~\cite{addor2017camels, kratzert2023caravan}, greenhouse-gas and carbon
fluxes~\cite{cheng2026agroflux,sun2026x,wang2026carbonglobe}, and air quality~\cite{betancourt2021aq}. Most of them report average scores
on a fixed test set. As a result, they tend to run into the same issues
we target, including scores saturate, performance drops over time, and models that look
good at monitored sites fail at new ones~\cite{nearing2024global}. EvoBench is an
evaluation layer that addresses these issues and can be applied across such
benchmarks. 

\paragraph{Large-scale hydrology datasets and models.}
CAMELS~\cite{addor2017camels, newman2015} provided catchment attributes and daily
data for 671 U.S.\ basins. Caravan~\cite{kratzert2023caravan} extended this idea
globally, CAMELSH~\cite{tran2025camelsh} provides hourly and daily data with
attributes for about 9{,}000 U.S.\ basins, and recent extensions add weather
nowcasts and forecasts to support operational, forecast-time
evaluation~\cite{caravanmultimet}. On these datasets, LSTM
models~\cite{kratzert2018} and the entity-aware LSTM
(EA-LSTM)~\cite{kratzert2019}, which uses static attributes to help a model
generalize across basins, became the standard baselines; transformers and other advanced machine learning models have been
tested more recently~\cite{tft2023,fan2026zerodiff,cheng2026hydrotft,yu2025physics,chen2022physics}. %Two recent directions are especially
%relevant to us. %For example, work on \emph{extremes} shows that deep rainfall--runoff
%models can be competitive on floods but remain sensitive to events outside the
%training range~\cite{frame2022extreme}, and large-scale models now predict floods
%even in \emph{ungauged} basins~\cite{nearing2024global}
In particular, several previous works have targeted extreme events and unseen scenarios~\cite{frame2022extreme,nearing2024global,chen2023meta} 
---exactly the regime and
spatial-transfer failures EvoBench isolates. General-purpose time-series
foundation models~\cite{ansari2024chronos, das2024timesfm} have also been 
applied to environmental series, which makes a diagnostic, evolving evaluation
even more useful. %All of these works report a single average NSE or KGE; our work
%is complementary, providing a new way to evaluate models rather than a new model.

\paragraph{Evaluation metrics.}
NSE~\cite{nash1970} and the Kling--Gupta efficiency~\cite{gupta2009, kling2012}
are the standard accuracy measures in hydrology. Flow-duration-curve
signatures~\cite{yilmaz2008} measure the bias in high flows and low flows
separately. We use these measures and add a skill score against a re-runnable
reference so that results stay comparable across rounds. Recent work has also explored using large language models to mimic human expert preferences in evaluation~\cite{cheng2026llm}.

\paragraph{Evolving and contamination-resistant benchmarks.}
A growing body of work keeps benchmarks fresh so they do not saturate or become
contaminated. Recent examples span modalities: collecting new data over time
(MMBench-Live~\cite{mmbenchlive}, LiveHouse-TS~\cite{livehousets}), scraping
freshly discovered problems to stay ahead of model cutoffs
(Live-kBench~\cite{livekbench2026}, SWE-MERA~\cite{swemera2025}), generating
harder items automatically~\cite{arenabencher}, and updating labels as evidence
improves~\cite{deepfact}. Recent surveys catalog these dynamic and
contamination-resistant designs but note there are still no agreed standards for
evaluating them~\cite{dynsurvey2025, contamsurvey2025}, and some argue live
benchmarks lose value once events are resolved~\cite{oracleproto}. Our setting is
different and offers a clear advantage. The test items come from the physical world, so there
is nothing to generate and nothing to leak, and old rounds remain valid because
the observed record only grows. %LiveHouse-TS~\cite{livehousets} is closest in
%spirit but is general-purpose; t
To our knowledge no evolving benchmark has been
built for environmental modeling. %, and none pairs the loop with a guardrail that
%checks physical plausibility.

\paragraph{Agents in evaluation.}
Recent benchmarks use language-model agents to build or check test
items~\cite{mmbenchlive, deepfact}. We use an agent in a narrow and auditable
way. Specifically, all the performance scores in our framework are computed deterministically.  %The deterministic code computes all scores. 
The agent only decides whether a
failure is a true model weakness or a data problem, and a model-independent
water-balance check backs up that decision.  The agent's outputs inform updates to the evaluation protocol across iterations but do not directly determine the final evaluation scores, avoiding the additional uncertainty that agent-driven scoring could introduce. %The agent's output is a label, never
%a score.

\section{EvoBench Instantiated for Streamflow}
\label{sec:benchmark}

%The five parts of EvoBench are designed to be general. Each needs only four
%things 
EvoBench requires multiple input components that most environmental-modeling tasks can provide: per-site time series of driver data, a
few static attributes describing each site, a physical or empirical model to use
as a reference, and a conservation law (such as a mass balance) to check whether
the observations are physically plausible. In this section, we explain each part
through the streamflow case study, \textbf{CAMELS-Evolve}. 
Fig.~\ref{fig:pipeline} illustrates the overall pipeline. Table~\ref{tab:generalize}
in the appendix shows how the same parts map to other domains.

\begin{figure*}[t]
\centering
\includegraphics[width=0.85\linewidth]{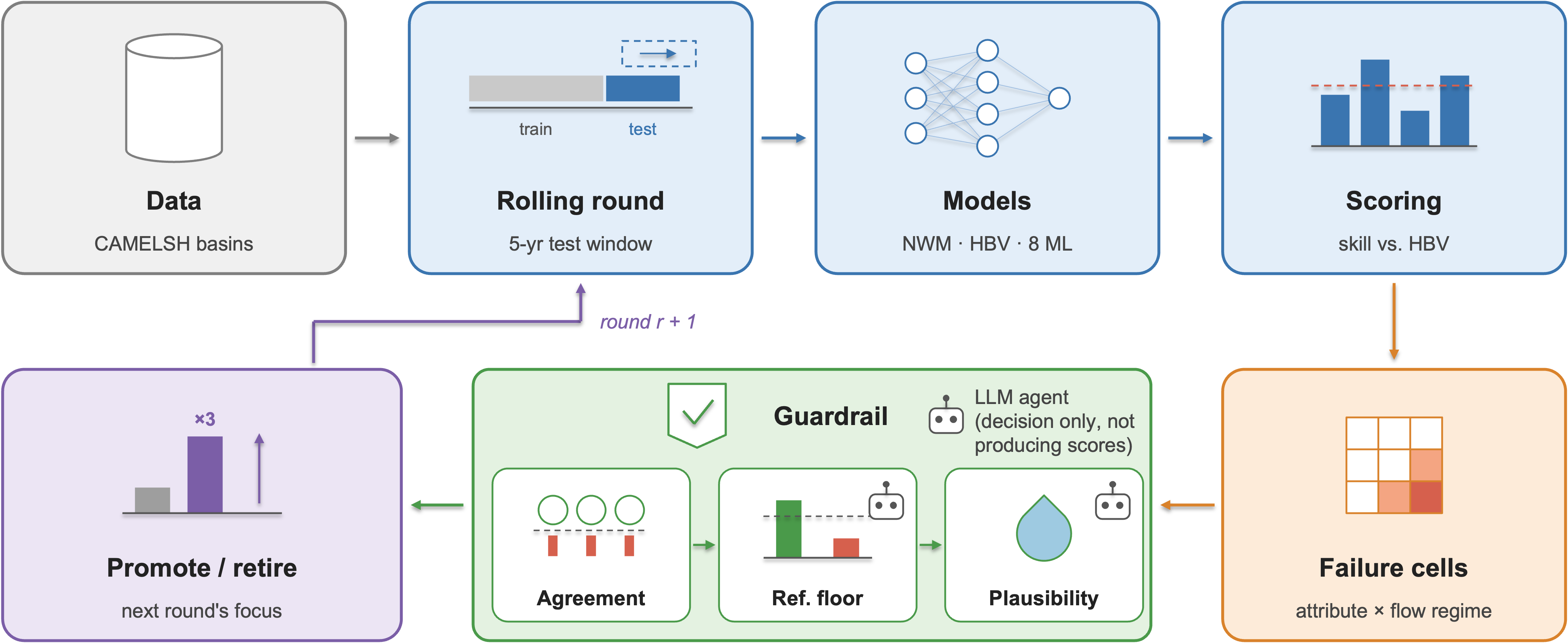}
\caption{Pipeline of the EvoBench framework.}
\label{fig:pipeline}
\end{figure*}

\subsection{Data}
We build the case study on CAMELSH~\cite{tran2025camelsh}, which contains 9{,}008
U.S.\ river basins with daily records from 1985 to 2024. For each basin and day
it provides 11 weather forcing variables, observed streamflow, and 44 static
catchment attributes (such as climate indices and soil and land-cover
descriptors), which are fixed in time. It also includes simulations from the National Water Model (NWM).
Because basins differ in how much observed streamflow is available, we use two
groups of basins: a strict \emph{core} group (162 basins with at least 90\% observation
coverage over 1985--2024) and a larger \emph{extended} observation group (1{,}139 basins with at
least 80\% coverage). %All attributes are fixed in time, which is why detecting
%changes within a basin (such as a new dam) is left to future work
%(Section~\ref{sec:limits}).

\textit{Dynamic input features.}
Each data-driven model reads a 180-day history of five daily meteorological
forcing variables, all derived from the NLDAS-2 reanalysis: precipitation
(\texttt{Rainf}), air temperature (\texttt{Tair}), downward shortwave radiation
(\texttt{SWdown}), downward longwave radiation (\texttt{LWdown}), and potential
evapotranspiration (\texttt{PotEvap}). These five variables drive the dominant
water-and-energy balance that governs runoff. Precipitation and temperature set
the input and the rain/snow partition; the two radiation terms and potential
evapotranspiration set the energy available to evaporate water and to melt snow.

\textit{Static catchment attributes.}
To let a single regional model adapt to each basin, we concatenate 18 static
attributes to the inputs, which capture the basin's  time-invariant climate, topography, soil, and land-cover characteristics. Specifically, they fall into four groups.
\emph{Climate}: mean precipitation, mean potential evapotranspiration,
aridity (their ratio), precipitation seasonality, the fraction of precipitation
falling as snow, and the frequency and duration of high- and low-precipitation
spells. \emph{Topography}: drainage area, mean elevation, and mean slope.
\emph{Soil}: sand, silt, and clay fractions, soil permeability, and the
baseflow index (the share of flow sustained by groundwater). \emph{Land cover}: forest fraction. These attributes are exactly the axes along which we
later define failure cells (Section~\ref{sec:benchmark}), so aridity, snow
fraction, area, baseflow index, and degree of regulation serve both as model
inputs and as diagnostic groupings.

\textit{Target and preprocessing.}
The prediction target is daily \emph{specific discharge} (mm\,day$^{-1}$), i.e.\
observed streamflow normalized by basin area, which makes basins of very
different size comparable. Forcings, attributes, and target are standardized to
zero mean and unit variance using statistics computed only from each round's
training window, so no future information leaks into the inputs. Days with
missing streamflow are excluded from training and scoring.

\subsection{Rolling 5-year evaluation rounds}
We evaluate models in \emph{rounds}. Each round covers five water years, i.e., the
most recent year plus the previous four years and the window moves forward by one
year between rounds. A model may use only data from before its round's window,
and it is then scored on the full five-year window. We use five years rather than
one because a single year could be dominated by year-specific conditions, e.g., whether it happened to be wet or dry,
while a five-year window is more stable and almost always contains both floods
and droughts. In our data, five-year rounds reduce the round-to-round variation  compared with one-year rounds (in scores by about 60\%). Because neighboring rounds
share four years of data, we treat a sequence of rounds as a trend rather than as
independent samples.

\subsection{Failure cells}
A \emph{failure cell} is a group of basins %(and, in the 2-D version, a flow
%regime) 
that share a common property. %The 1-D version 
Specifically, we group basins by five attributes,
each linked to a known weakness of data-driven models: aridity (dry basins and
low flow), snow fraction (snowmelt), basin area (long memory and routing),
baseflow index (flashy response), and degree of regulation (dams and human
control). The 2-D version  additionally splits each basin's record by flow
regime---low, medium, high, and peak---using that basin's own flow-duration
curve. The 2-D view is important because a single average score can hide large errors that occur only in specific regimes. 
When analyzing predictive bias, 
opposite errors in different regimes can cancel out. For example, a model may over-predict low flow and under-predict peaks at the same time, which looks only ``average'' overall
but is clearly wrong once the two regimes are separated.

%opposite errors that cancel out. For example, a model may over-predict low flow

\subsection{The evolving loop and its guardrail}
\label{sec:loop}
Each round, the loop finds the failure cells where even the best model does poorly
and \emph{promotes} them, i.e., keeping them as the focus for the next round. The loop  also  
\emph{retires} cells that a model has solved (median NSE at least 0.5). Over time,
attention moves toward the problems that remain unsolved. However, a loop that simply chased
the largest errors would drift toward meaningless cases that cannot be possibly predicted by any models given the available data, such as dammed basins or
bad gauges. Hence, the promotion of failure cells must pass a \textit{guardrail} with three checks:
\begin{enumerate}[leftmargin=1.4em]
\item \emph{Agreement across independent models.} We first require the failure to be shared across multiple models, rather than being specific to a single model, to ensure that it represents a meaningful limitation worth highlighting in the benchmark. %A cell is a candidate only if several of the independent baselines do poorly on it---ideally spanning different families, such as the physical NWM, the conceptual HBV, and the data-driven networks. 
If a single model fails while the others succeed, that
is a weakness of that model, not a capability gap worth promoting for the field. 
In our implementation, %practice 
we flag a cell only when at least two independent data-driven models  fall below
a preset skill threshold, which filters out the artifact of any one architecture or training
run.

\item \emph{A reference floor.} We then ask whether the target is even
\emph{learnable} from the available data, by comparing against the
per-basin-calibrated HBV reference and the physical NWM. If HBV or NWM predicts the cell acceptably while the learned models fail, the signal clearly exists in the data, so the failure is a genuine learned-model capability gap and should be promoted. If instead \emph{every} model---including the calibrated HBV---fails, the cell is more likely limited by its data than by the models, and promoting it would chase an unlearnable target. To make this judgment, an LLM agent reads a short, structured evidence summary for the cell---the per-model skill
scores, the HBV reference floor, the degree of regulation, and other static
attributes---and labels it \emph{capability gap}, \emph{regulation effect},
\emph{data problem}, or \emph{needs external information}. %A deterministic rule over the same evidence (multi-model agreement, the reference floor, and a regulation threshold) serves as a transparent fallback; it agreed with the agent on the large majority of worst cells, and where they differed the agent was the more conservative, declining to call a capability gap whenever no model-independent signal showed the target was predictable.

\item \emph{Physical plausibility.} Finally, a \emph{model-independent} check
tests whether the observations themselves are trustworthy, using a conservation
law rather than any prediction. Over a long record runoff cannot exceed
precipitation, so the runoff ratio, i.e., total runoff divided by total
precipitation, must not exceed one. A value above one %signals a broken target,
%most 
often indicates a wrong drainage area, an inter-basin transfer, or a gauging error,
and thus the basin will be removed as a data artifact. %This decisive test is fully deterministic (runoff ratio $>1 \Rightarrow$ exclude). 
Even if the data meet the runoff ratio requirement, it could be flagged % signal  flagged but is not
%physically impossible---
if we observe that an runoff ratio very close to one, an
implausibly high ratio for an arid basin, negative flows, or a long constant-value
segment. %, or low coverage. 
These flags mark a basin as suspect but cannot by
themselves say whether the data are wrong.  For example, a long constant-value
segment could be either a stuck sensor (bad data) or a managed, constant
reservoir release (a real flow). Only these ambiguous cases are referred to a verification agent, which uses the basin's regulation context to make the final decision. In the previous example,  a flat segment in a heavily regulated basin is read as a managed release and kept, whereas the same pattern in an unregulated basin is treated as a faulty sensor. %Which case a basin falls into is set by the fixed rules, not by the agent; the agent onlyresolves case~(iii), and its label never enters a score. 
% This check also settles
% basins the reference-floor step could not: in one basin where every model,
% including HBV, failed, a runoff ratio of $1.6$ showed the fault lay in the
% observations rather than the models, so the cell was labeled a data problem
% rather than a capability gap.
% Genuinely ambiguous cases---for instance a long constant-value segment that could be either a stuck
% sensor or a managed, constant reservoir release---are instead referred to the
% agent, which weighs the basin's regulation context before deciding. In one case
% this check settled a basin where every model including HBV failed: a runoff ratio
% of $1.6$ showed the fault lay in the observations, not the models.
\end{enumerate}
Only a cell that passes all three checks, i.e., a failure shared across models, shown
to be learnable by the reference, and backed by physically plausible data, is
promoted. A key consequence is that the guardrail does not blindly promote the
flood regime. For example, when every model fails on peak flows and the reference gives no
signal, the cell is labeled a data problem. This could be because high-flow observations are
themselves uncertain (e.g., rating curves extrapolated beyond the measured range). %: they come from rating curves extrapolated beyond the measured range. Throughout, 
It is noteworthy that the agents used in our framework only make decisions and do not produce performance scores. All performance scores are computed by a deterministic process,  which mitigates the impact of LLM stochasticity and improves the reproducibility of the benchmark results.  %the deterministic code computes every score; the
% agent only produces labels, which never enter any score.

\paragraph{How much each cell is promoted, and its effect on the next round.}
Promotion is graded rather than binary: a cell is weighted by how far it sits
below the ``solved'' line. Let $\tau$ be the retire threshold (we use a median
NSE of $0.5$) and $s_c$ the cell's current skill; its shortfall is
$g_c = \max(0,\, \tau - s_c)$. The cell's weight for the next round is
\begin{equation}
w_c \;=\; 1 + \beta\, g_c/\tau , \qquad w_c \in [1, w_{\max}],
\label{eq:promote}
\end{equation}
where we set $\beta$ and the cap $w_{\max}$ so that the worst, persistently open
cells reach about $3\times$ the weight of a just-solved one; retired cells keep
$w_c = 1$ and move to a regression set that is still scored but no longer
emphasized. This weight feeds the next round in two concrete ways. First, in
\emph{scoring}: the round's headline skill is a weighted average over cells,
$\sum_c w_c s_c / \sum_c w_c$, so a cell that stays far below $\tau$ counts up to
$3\times$ as much as an easy one, and the leaderboard increasingly rewards
progress on the current frontier. Second, in \emph{sampling}: basins belonging to
promoted cells are over-sampled into the next round's evaluation set---and
participants may up-weight them during training---in proportion to $w_c$, which
concentrates both data and effort where models currently fail. Because every
per-cell skill is measured against the same re-runnable reference
(Eq.~\ref{eq:skill}), we also report the \emph{unweighted} per-cell scores so
that cross-round trends stay interpretable; the weighted headline reflects only
the current round's emphasis. In our experiments the sampling side is
instantiated with a $3\times$ up-weighting of promoted-cell basins
(Section~\ref{sec:exp}).

\subsection{Comparable scores across time}
Some years are much harder to predict than others, so raw accuracy cannot be
compared directly across rounds. Instead we report a \emph{skill} score relative
to a reference model, computed in the same round:
\begin{equation}
\mathrm{skill} = 1 - \mathrm{MSE}(\text{model}) / \mathrm{MSE}(\text{reference}).
\label{eq:skill}
\end{equation}

%The performance measure in Eq.~\ref{eq:skill} can also be replaced by other metrics (e.g., NSE, KGE). 
If a specific round is hard, both the model's error and the reference's error grow
together, so their ratio stays stable. The reference must be runnable in any
period. %The NWM simulations end in early 2023, so NWM cannot play this role for
%future rounds. We therefore 
In this work, we use HBV~\cite{seibert2012hbv}, a simple %10-parameter
process-based hydrology model that we calibrate once per basin and can then run on any input.
HBV is meant to serve only as a modest baseline, not a strong competitor. %; 
%In contrast, NWM is included as a second, stronger reference where it is available.

\subsection{Prediction at unmonitored sites}
%Standard evaluation holds out only time, so the model still sees every test basind uring training. 
Many existing works holds out only time in evaluation. 
To measure true spatial generalization, we group basins into six
clusters by their attributes and hold out one whole cluster at a time, training on
the rest and predicting the held-out basins. We report the \emph{spatial gap}, the
drop in NSE from gauged to ungauged prediction. For the reference, HBV's drop from
local calibration to using neighbors' parameters gives a baseline that learned
models should beat.
\section{Experiments}
\label{sec:exp}

\subsection{Setup}
We compare ten models. Two are process-based models: the operational {NWM}
and the {HBV} reference. Eight are data-driven baselines. Four are models
commonly used in hydrology: {LSTM},
{EA-LSTM}~\cite{kratzert2019towards}, a \textbf{GRU} (lighter gated recurrent model), and
a small encoder-only {Transformer} (2 layers, model width 64, about 69k
parameters). The other four are popular general-purpose time-series
architectures: a \textbf{TCN} (dilated causal convolutions), \textbf{DLinear}
(trend--seasonal decomposition followed by linear maps over time)~\cite{zeng2023transformers},
\textbf{TSMixer} (all-MLP mixing across time and features)~\cite{chen2023tsmixer}, and a simplified
\textbf{PatchTST} (a 2-layer Transformer encoder over 30-day patches)~\cite{nie2022time}. The training data covers all the data before the testing window in the current round. Each data-driven model reads the input of a 180-day history of five weather
variables plus 18 static attributes and predicts daily streamflow. Every
data-driven model is trained for one epoch, i.e., one expected pass over all of
its training samples, so the number of iterations grows with the size of the
training set; full training details are in Appendix~\ref{app:repro}. We report results on the \emph{core}
group (162 basins) and the \emph{extended} group (1{,}139
basins). We use five metrics: NSE and KGE$'$ for overall accuracy, log-NSE for
low flows, peak-day percent bias for floods, and the skill score of
Eq.~\eqref{eq:skill} against HBV.

\subsection{Fixed-split test}
\subsubsection{Model performance on fixed split}
\paragraph{Setting.} This comparison uses a single fixed temporal split rather
than the sliding rounds of Section~\ref{sec:loop}. All data-driven models are
trained on all data before the test window (water years 1986--2010) and
evaluated on 2011--2022, a held-out future period, matching the protocol of the
sliding rounds. The HBV reference is calibrated per basin on the same
pre-test record (water years 1986--2010). NWM is scored from NOAA's NWM v3.0
retrospective simulation over the test period; it is an external operational
benchmark whose parameters were calibrated by its developers rather than under
our protocol. %, and its calibration period may overlap our test years, which
%would favor NWM.
%The finding that no data-driven model surpasses NWM is therefore conservative.
%Each data-driven model is trained once (a single seed, no ensembling)
%with the shared setup of Appendix~\ref{app:repro}, and e
Every reported value is a
median over basins.

Table~\ref{tab:head} shows the comparison. The ranking changes with
the metric. On overall accuracy, the LSTM (NSE 0.674) and GRU (0.665) surpass
NWM (0.644), and the LSTM also has the best KGE$'$ (0.711) and the best low-flow
skill (log-NSE 0.595).
Here note that NWM is an external operational
benchmark whose %parameters were calibrated by its developers rather than under
%our protocol. %, and its
calibration period may overlap our test years, which would favor NWM. %, so this lead of the data-driven models is, if anything, understated.
NWM remains by far the best at floods: every data-driven model
under-predicts peak days by 27--44\%, while NWM is at $-13\%$. The EA-LSTM
(0.633) comes close to NWM on NSE, while the GRU, despite its high NSE, is weaker
at low flows (log-NSE 0.545). Among data-driven time-series architectures, only the TCN (0.595) is competitive
with the recurrent models; DLinear, PatchTST, and TSMixer fall below the HBV
reference on NSE and have strongly negative log-NSE, i.e., they largely fail to
learn low flows. The  Transformer lies in between on the 162-basin core group
(NSE 0.558). At the larger scale of the extended group
(Table~\ref{tab:cov80}), most models score lower, because this group admits
harder, lower-coverage, and more regulated basins. The LSTM (0.637), GRU
(0.630), and EA-LSTM (0.586) all surpass NWM (0.562), by a wider margin than on
the core group, and the LSTM is again the best at low flows (log-NSE 0.606). NWM
keeps the best KGE$'$ (0.664) and peak-day bias ($-14\%$, against $-29\%$ to
$-45\%$ for the data-driven models). A single-number leaderboard would hide all
of these differences: the best model depends on whether one asks about average
skill, low flows, or floods.

\paragraph{Comparison with prior CAMELS results.} Our absolute NSE values are
lower than the strongest LSTM results reported on the standard 531-basin CAMELS
benchmark, where well-tuned, ensembled models reach a median NSE of roughly
0.74--0.76~\cite{kratzert2019, feng2020}. Several deliberate choices explain the
difference. We train a single model %per baseline (one seed, no ensembling), use
using a
reduced input set (five forcings and 18 attributes), and maintain the same %hold a fixed, modest
training budget %(one epoch) %identical
across all ten models, so that differences reflect the
models rather than tuning effort. We also evaluate on CAMELSH with a different
temporal split, and our extended group deliberately includes low-coverage and
heavily regulated basins that the 531-basin benchmark excludes, both of which
lower the median. Also note that our aim is to expose \emph{relative}, regime-specific
behavior rather than to maximize absolute skill. %, these controlled baselines
%suffice; s
Stronger absolute performance could be produced by %would follow from 
model-specific tuning, ensembling, and augmentation of input variables. %, %richer inputs,longer training, and per-model tuning.

\begin{table}[t]
\centering
\caption{Head-to-head on the core group (162 basins, WY2011--2022), median over
basins. The best value in each column is in \textbf{bold}; the order changes by
metric.}
\label{tab:head}
\small
\begin{tabular}{lccccc}
\toprule
Model & NSE & KGE$'$ & log-NSE & peak\%bias & skill/HBV \\
\midrule
LSTM        & \textbf{0.674} & \textbf{0.711} & \textbf{0.595} & $-$32.2 & $+$0.368 \\
GRU         & 0.665 & 0.688 & 0.545 & $-$30.6 & \textbf{$+$0.379} \\
NWM         & 0.644 & 0.694 & 0.563 & \textbf{$-$13.0} & $+$0.332 \\
EA-LSTM     & 0.633 & 0.689 & 0.521 & $-$28.9 & $+$0.337 \\
TCN         & 0.595 & 0.619 & 0.465 & $-$42.1 & $+$0.273 \\
Transformer & 0.558 & 0.498 & $-$0.910 & $-$27.0 & $+$0.170 \\
HBV (ref)   & 0.482 & 0.609 & $-$0.168 & $-$37.1 & 0.000 \\
PatchTST    & 0.465 & 0.471 & $-$2.018 & $-$43.9 & $+$0.144 \\
DLinear     & 0.439 & 0.485 & $-$0.457 & $-$41.1 & $-$0.130 \\
TSMixer     & 0.323 & 0.307 & $-$1.172 & $-$38.1 & $-$0.232 \\
\bottomrule
\end{tabular}
\end{table}

\begin{table}[t]
\centering
\caption{Head-to-head at larger scale (extended, 1{,}139 basins), median over
basins. The LSTM, GRU, and EA-LSTM surpass NWM on NSE, while NWM keeps the best
KGE$'$ and flood (peak-day) bias.}
\label{tab:cov80}
\small
\begin{tabular}{lccccc}
\toprule
Model & NSE & KGE$'$ & log-NSE & peak\%bias & skill/HBV \\
\midrule
LSTM        & \textbf{0.637} & 0.653 & \textbf{0.606} & $-$31.2 & \textbf{$+$0.391} \\
GRU         & 0.630 & 0.661 & 0.477 & $-$29.3 & $+$0.373 \\
EA-LSTM     & 0.586 & 0.595 & 0.502 & $-$34.4 & $+$0.314 \\
NWM         & 0.562 & \textbf{0.664} & 0.468 & \textbf{$-$13.6} & $+$0.261 \\
TCN         & 0.525 & 0.567 & 0.325 & $-$44.2 & $+$0.208 \\
Transformer & 0.517 & 0.529 & $-$0.337 & $-$31.2 & $+$0.186 \\
PatchTST    & 0.465 & 0.424 & $-$2.447 & $-$35.7 & $+$0.151 \\
HBV (ref)   & 0.406 & 0.547 & $-$0.010 & $-$41.9 & 0.000 \\
DLinear     & 0.316 & 0.369 & $-$0.566 & $-$44.7 & $-$0.237 \\
TSMixer     & 0.203 & 0.269 & $-$1.445 & $-$33.4 & $-$0.337 \\
\bottomrule
\end{tabular}
\end{table}

\subsubsection{Diagnostic tests}
These tests reuse the fixed-split predictions above (trained to 2010, tested on
water years 2011--2022) on the core group, and probe specific regimes and the
spatial dimension.
\paragraph{Floods.} When we score only peak days (observed flow above the 99th
percentile), all data-driven models under-predict badly (LSTM $-34\%$, EA-LSTM
$-32\%$, Transformer $-27\%$), while NWM stays at $-13\%$. HBV is the worst of
these ($-41\%$). Floods are a shared weak point, and the one regime where the
data-driven models clearly trail NWM.
\paragraph{Dry versus wet years.} Splitting each basin's record into dry, normal,
and wet years shows that every model does worse in dry years. The Transformer has
the largest dry-to-wet gap (NSE 0.41 to 0.55), meaning it generalizes worst across
climate conditions.
\paragraph{Unmonitored basins.} Table~\ref{tab:pub} reports the leave-cluster-out
test: basins are grouped into six clusters by $k$-means on standardized
attributes, and each cluster is held out in turn while the model is trained on
the other five, so the held-out basins are never seen in training. Each fold
model is trained over its own training basins. %, the same budget
%rule as above. %The result is one a gauged-only leaderboard would miss: a
Although EA-LSTM is worse on monitored basins, its drop to unmonitored basins is about 45\%
smaller than the LSTM's (0.105 vs 0.191), and it becomes the best data-driven
model once ungauged (NSE 0.529). This matches the design of
the EA-LSTM, whose use of static attributes is meant to help it transfer across
locations. Conversely, the LSTM, the best model on monitored basins, loses the
most among the strong models (gap 0.191), followed by the TCN (0.175) and GRU
(0.165). All data-driven models fall well
below their gauged scores, and only the EA-LSTM, GRU, and LSTM stay clearly
above the HBV regional reference (0.416); the Transformer and PatchTST, with the
largest gaps (about 0.25), fall below it. The very small or negative gaps of
DLinear and TSMixer reflect weak gauged skill rather than good transfer.

\begin{table}[t]
\centering
\caption{Prediction at unmonitored basins (core), using leave-6-cluster-out
testing. The EA-LSTM transfers best despite a lower score on monitored basins.}
\label{tab:pub}
\small
\begin{tabular}{lccc}
\toprule
Model & NSE gauged & NSE ungauged & spatial gap \\
\midrule
LSTM     & \textbf{0.674} & 0.484 & 0.191 \\
GRU      & 0.665 & 0.499 & 0.165 \\
EA-LSTM  & 0.633 & \textbf{0.529} & 0.105 \\
TCN      & 0.595 & 0.420 & 0.175 \\
Transformer & 0.558 & 0.306 & 0.252 \\
PatchTST & 0.465 & 0.216 & 0.250 \\
DLinear  & 0.439 & 0.375 & 0.064 \\
TSMixer  & 0.323 & 0.363 & $-$0.039 \\
HBV (local$\to$regional) & 0.482 & 0.416 & 0.074 \\
\bottomrule
\end{tabular}
\end{table}

\subsubsection{Failure cells}
\begin{figure}[t]
\centering
\includegraphics[width=0.98\linewidth]{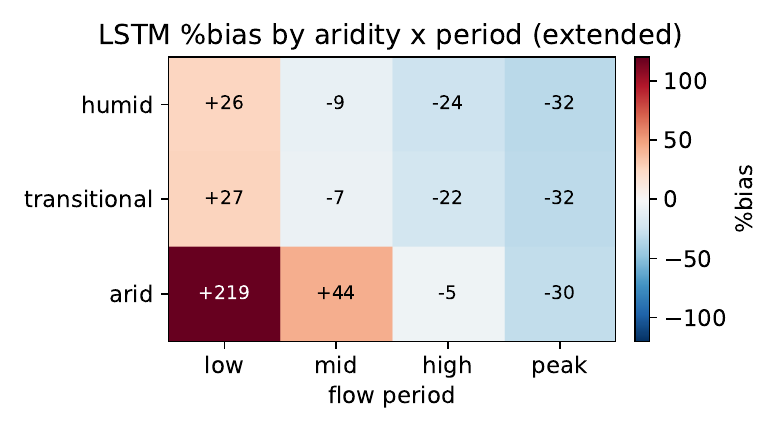}
\caption{2-D failure grid: LSTM percent bias by aridity and flow regime (extended).
Low flows are over-predicted (red, left) and peaks are under-predicted (blue,
right). The two errors cancel in a single NSE.}
\label{fig:grid2d}
\end{figure}

Here we use the fixed-split predictions on the extended group (1{,}139 basins),
which gives enough basins per cell for stable estimates. The 1-D grid
consistently ranks the hardest basin types across models: at the
larger scale, snow-dominated basins (n$=$109; LSTM NSE 0.47), arid basins
(n$=$492; 0.50), and baseflow-dominated basins (0.57). Heavily regulated basins
are hard for NWM (0.49 vs.\ 0.60 on natural basins) and for the EA-LSTM and
Transformer, but barely for the LSTM (0.63 vs.\ 0.64). The 2-D grid
(Figure~\ref{fig:grid2d}) explains why the LSTM looks only average on arid basins:
it makes two opposite errors that cancel in a single score. It over-predicts low
flow by about 219\% and under-predicts peaks by about 30\%. The only cell where
the LSTM is worse than the HBV reference is arid low flow (skill relative to HBV
$-0.31$). Because the simple reference succeeds there, the signal exists in the
data, so this cell passes the reference-floor check of the guardrail as a
genuine model capability gap. On the peak cells, in contrast, the LSTM beats HBV
(skill about $+0.3$) yet still under-predicts peaks by 30--37\%, while NWM is at
about $-14\%$; since the physical reference captures floods much better from the
same weather, the flood shortfall is also a learnable capability gap rather than
a pure data limitation, although high-flow observations, which come from rating
curves extrapolated beyond the measured range, add uncertainty for every model.
The 2-D grid also reveals differences between models that an average
score hides: on arid low flow the LSTM over-predicts ($+219\%$) while the
Transformer under-predicts ($-62\%$). The two models fail in opposite directions.

\subsection{Evolving Tests}
\subsubsection{How performance changes over time}
\begin{figure}[t]
\centering
\includegraphics[width=0.98\linewidth]{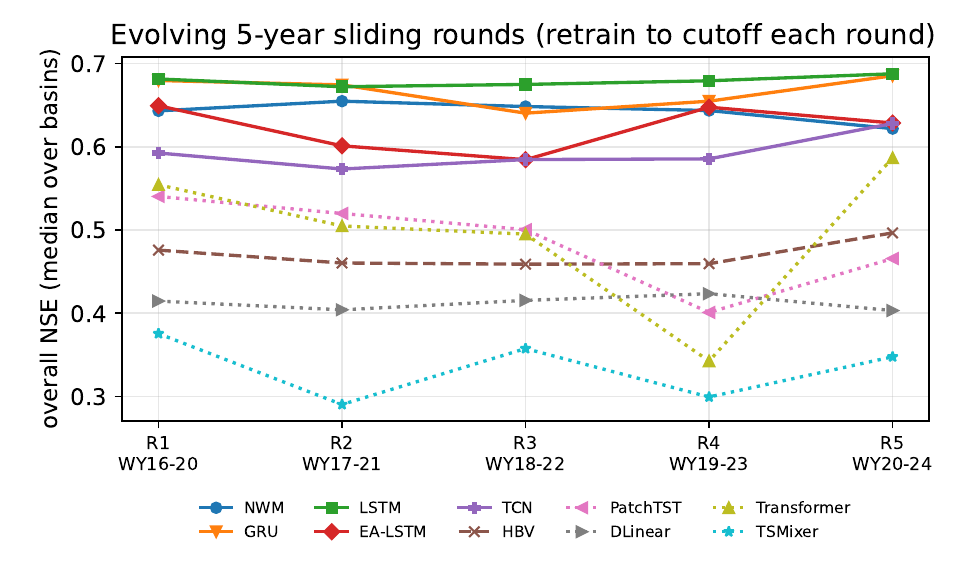}
\caption{Five-year sliding rounds (core): overall NSE for each model over time.
The LSTM and GRU stay above NWM in nearly every round, while weaker models swing
between rounds; HBV is a flat reference. NWM's rounds 4--5 cover only part of
the window (record ends January 2023).}
\label{fig:evolving}
\end{figure}

We run the 5-year window forward for five rounds, retraining every data-driven
model from scratch to each round's cutoff (Figure~\ref{fig:evolving},
Table~\ref{tab:rounds}). The rounds test water years 2016--2020, 2017--2021,
\ldots, 2020--2024. In each round a model is trained on all data before its
window (round~1 on $\le$2015, round~5 on $\le$2019), and we evaluate on both the
core and the extended group, with HBV recalibrated to each round's cutoff as the
reference. We see four things.
First, the ranking is stable over time: the LSTM surpasses NWM in every round
on both groups and has the best overall NSE in all rounds but one (core round~2,
where the GRU is marginally ahead), and the GRU surpasses NWM in every round
except round~3 on the core group. The EA-LSTM and TCN follow, while the Transformer and
the other general-purpose architectures stay near or below HBV.
Second, the strongest models are stable across rounds: the LSTM varies by less
than 0.02 NSE over the five rounds on both groups, even though the rounds
include harder recent years. Less stable models still swing between rounds:
the EA-LSTM dips in the middle rounds on the core group (0.649 in round~1 to
0.585 in round~3), and the Transformer collapses in single rounds (0.343 in
round~4 on the core group, 0.154 in round~5 on the extended group). A fixed split
would average such swings away. HBV stays a flat, low baseline and keeps the skill
scores comparable. Third, arid and snow basins remain the frontier, and they
show the retire mechanism at work. On the core group, no data-driven model
reaches the retirement threshold (median NSE 0.5) on arid basins in any round,
and NWM reaches it only once (round~2). On the extended group, the LSTM reaches
it on arid basins in rounds~1 and~4 (0.502 and 0.506) and on snow basins in
rounds~1, 4, and~5 (0.525, 0.504, 0.559); because these cells sit right at the
threshold, they alternate between retired and open across rounds, which is why
retired cells stay in a regression set rather than being dropped. Averaged over
rounds, the LSTM matches NWM on core arid basins (0.410 vs.\ 0.409) and exceeds
it on extended arid (0.480 vs.\ 0.381) and snow basins (0.505 vs.\ 0.265); the
core group has too few snow basins (n$=$6) for stable estimates. Fourth, the NWM retrospective ends in January 2023, so
in rounds~4 and~5 NWM is scored only on the early part of the window. Its drop
in round~5 (extended overall 0.530, snow $-$0.099) therefore should not be read
as a real decline. This is why the benchmark uses HBV, which can be rerun on any
new window, as its reference.

\begin{table}[t]
\centering
\caption{Five-year sliding rounds: median NSE averaged over the five rounds, for
the core group (162 basins) and the extended group (1{,}139 basins). Every
data-driven model is retrained to each round's cutoff. Best per column in
\textbf{bold}.}
\label{tab:rounds}
\small
\begin{tabular}{lccccc}
\toprule
& \multicolumn{2}{c}{Core} & \multicolumn{3}{c}{Extended} \\
\cmidrule(lr){2-3}\cmidrule(lr){4-6}
Model & overall & arid & overall & arid & snow \\
\midrule
LSTM        & \textbf{0.679} & \textbf{0.410} & \textbf{0.644} & \textbf{0.480} & \textbf{0.505} \\
GRU         & 0.667 & 0.359 & 0.632 & 0.476 & 0.469 \\
NWM         & 0.642 & 0.409 & 0.585 & 0.381 & 0.265 \\
EA-LSTM     & 0.622 & 0.367 & 0.610 & 0.418 & 0.375 \\
TCN         & 0.593 & 0.264 & 0.517 & 0.329 & 0.294 \\
Transformer & 0.497 & 0.052 & 0.397 & 0.184 & 0.048 \\
PatchTST    & 0.486 & 0.144 & 0.478 & 0.208 & $-$0.010 \\
HBV (ref)   & 0.470 & 0.302 & 0.413 & 0.285 & 0.137 \\
DLinear     & 0.412 & $-$0.410 & 0.323 & $-$0.099 & $-$0.600 \\
TSMixer     & 0.334 & $-$0.509 & 0.246 & $-$0.136 & $-$0.312 \\
\bottomrule
\end{tabular}
\end{table}

\subsubsection{Enhancing model based on benchmark feedback. }
\begin{figure}[t]
\centering
\includegraphics[width=0.9\linewidth]{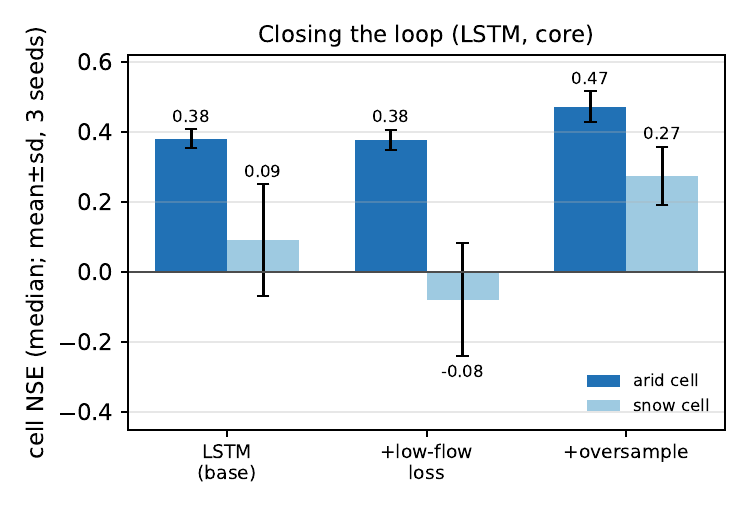}
\caption{Fixing a diagnosed weakness (LSTM, core, fixed split): arid- and
snow-cell NSE, mean $\pm$ standard deviation over three seeds. The low-flow loss
does not help, while over-sampling lifts both the arid and the snow cell.}
\label{fig:closing}
\end{figure}

Finally, we test whether the benchmark's findings can be acted on, using the
fixed split on the core group. The benchmark flags that the LSTM over-predicts
low flow in arid basins (by about $+63\%$ on arid low-flow days) and that arid
and snow basins remain its weakest cells. In response,
we apply two simple changes to training: we give low-flow days three times more
weight in the loss, and we additionally sample arid and snow basins three times
more often. Because single runs are noisy, each configuration is trained with
three seeds (Figure~\ref{fig:closing}). The low-flow loss does not help: the
arid-cell NSE is unchanged (0.381 to 0.378), the arid low-flow bias barely moves
($+63\%$ to $+61\%$), and the snow cell gets worse (0.09 to $-0.08$).
Over-sampling, in contrast, lifts the arid cell from 0.381 to 0.473, more than
twice the seed spread (standard deviation 0.03--0.04), and the snow cell from 0.09
to 0.28, with no loss in overall NSE (0.668 in both cases). The arid cell thus
moves close to the retirement threshold, crossing it for one of the three seeds,
but it stays open on average and is carried to the next round. The experiment
also shows that the benchmark can tell which of two plausible fixes actually
addresses a diagnosed weakness, and why it reports seed spread: the per-seed arid
low-flow bias ranges from $+8\%$ to $+130\%$ for the same configuration, so bias
changes from a single run would be unreliable.

\subsubsection{Diagnosis, intervention, and memory in evolving tests.}
We now ask whether this loop generalizes. For each data-driven baseline we run
three conditions over the five sliding rounds on the core group. \emph{(1) Baseline}: a fresh model
each round, no intervention. \emph{(2) Targeted}: a fresh model each round, but
intervened on the previous round's promoted failure cells (a low-flow-weighted
loss plus over-sampling of basins in those cells). \emph{(3) Targeted+memory}:
the same intervention, but the model is carried across rounds (warm start) with
cheap incremental updates (20\% of a one-epoch training per round, about
1{,}300--1{,}500 iterations instead of 6{,}400--7{,}300). Round~1
is identical for all three, since there is no previous round to diagnose.
Tables~\ref{tab:three} and~\ref{tab:threearid} report the mean over the five
rounds. Each cell is a single run, so given the seed spread measured above
(about 0.01 for overall and 0.03--0.04 for arid NSE), we only read differences larger
than these as real. (The baseline column differs slightly from
Table~\ref{tab:rounds} because here input normalization is fixed from the
round-1 window for all rounds, so that a warm-started model sees consistent
inputs.)

Three patterns are observed. First, the targeted intervention helps most models
on the cell it aims at. On arid basins it clearly raises the GRU (0.360 to
0.442, and 0.469 with memory) and the EA-LSTM (0.361 to 0.401, and 0.422 with
memory), and with memory also the TCN (0.240 to 0.294); it lifts the linear and
all-MLP models from very low levels (TSMixer $-0.50$ to $-0.15$, DLinear $-0.41$
to $-0.27$ with memory). The LSTM, which already has the best arid score
(0.457), does not gain further, and the Transformer does not benefit at all.
Second, cross-round memory gives the best overall score for five of the eight
models while using about a fifth of the compute per round: the LSTM (0.680 to
0.690), GRU (0.668 to 0.679), EA-LSTM (0.629 to 0.637), TSMixer (0.355 to
0.375), and PatchTST (0.465 to 0.512). For the three recurrent models this gain
is small, close to the seed spread, so memory mainly matches full retraining at
a fraction of the cost. Third, the interventions do not help every family: they
lower the Transformer's overall score (0.503 to 0.484), and memory slightly
lowers the TCN (0.579 to 0.569). In short, the loop ``diagnose $\to$ intervene $\to$ update
with memory'' helps most baseline models on the evolving benchmark, but whether a
given intervention helps also depends on the model family. %, which is itself a findingthat a fixed leaderboard cannot surface.

\begin{table}[t]
\centering
\caption{Overall NSE (mean over five sliding rounds, core) for each baseline
under the three conditions. Best per row in \textbf{bold}.}
\label{tab:three}
\small
\begin{tabular}{lccc}
\toprule
Model & (1) baseline & (2) targeted & (3) targeted+memory \\
\midrule
LSTM        & 0.680 & 0.682 & \textbf{0.690} \\
EA-LSTM     & 0.629 & 0.625 & \textbf{0.637} \\
GRU         & 0.668 & 0.660 & \textbf{0.679} \\
Transformer & \textbf{0.503} & 0.484 & 0.484 \\
TCN         & \textbf{0.579} & 0.575 & 0.569 \\
DLinear     & \textbf{0.412} & 0.407 & 0.408 \\
TSMixer     & 0.355 & 0.356 & \textbf{0.375} \\
PatchTST    & 0.465 & 0.502 & \textbf{0.512} \\
\bottomrule
\end{tabular}
\end{table}

\begin{table}[t]
\centering
\caption{Arid-cell NSE (mean over five rounds, core), where the intervention is
aimed. The intervention helps most models, most clearly the GRU and EA-LSTM;
the LSTM, already the strongest on arid basins, and the Transformer do not gain.}
\label{tab:threearid}
\small
\begin{tabular}{lccc}
\toprule
Model & (1) baseline & (2) targeted & (3) targeted+memory \\
\midrule
LSTM        & \textbf{0.457} & 0.451 & 0.446 \\
EA-LSTM     & 0.361 & 0.401 & \textbf{0.422} \\
GRU         & 0.360 & 0.442 & \textbf{0.469} \\
Transformer & \textbf{0.043} & 0.038 & $-$0.017 \\
TCN         & 0.240 & 0.249 & \textbf{0.294} \\
DLinear     & $-$0.408 & $-$0.369 & \textbf{$-$0.274} \\
TSMixer     & $-$0.500 & $-$0.251 & \textbf{$-$0.147} \\
PatchTST    & 0.116 & 0.087 & \textbf{0.125} \\
\bottomrule
\end{tabular}
\end{table}

\section{Conclusion}
\label{sec:conclusion}

We presented \textbf{EvoBench}, a benchmark framework for environmental modeling
that is both diagnostic and evolving, and we validated it through a streamflow
case study (\textbf{CAMELS-Evolve}) on CAMELSH. %Instead of a single average score, EvoBench reports skill against a re-runnable reference so that rounds are comparable, 
EvoBench breaks performance into failure cells by site attribute and regime, measures prediction at unmonitored sites, slides a time window forward to track
which failures remain, and uses a small agent with a physical consistency check
to tell true model weaknesses apart from data problems. %These parts are general and map onto other per-site environmental tasks (Table~\ref{tab:generalize}).

Across the base models in our tests, we found that no model is best at everything. %: NWM has thebest average accuracy, the LSTM is best at low flows, the EA-LSTM generalizes best to unmonitored basins, the Transformer is strong only at scale, and 
Also all data-driven models under-predict floods. %
A single number would hide all of these differences. %We also showed the findings are useful: a targeted fix removed the arid low-flow weakness the benchmark identified, while the same fix did not help snow basins, whose weakness has a different cause. 
Finally, evolving the benchmark by sliding the window
forward showed varying model performance, %that data-driven models get worse in recent, harder years, 
a sign
of non-stationarity that a fixed split hides, while snow and arid basins remained
the hardest cases, with the best models only beginning to retire them in some
rounds. %We believe t
This shift---from a single ranking on a static dataset to a
moving map of model limitations---is useful for environmental ML benchmarks in
general, where the real environmental ecosystems keep supplying new benchmark data.

\section{Limitations}
\label{sec:limits}
%Our study has several limits. First, we used modest training budgets (one seed,no ensembles), and GPU training is not perfectly reproducible, so the exactscores carry about $\pm0.02$ of noise; the trends, not the exact numbers, are the point. Second, the weather forcing is reanalysis data, which already incorporates observed weather. Our evaluation is therefore a hindcast that tests the rainfall-to-runoff mapping, not true forecasting; a clean forecasting setup would need forecast weather inputs. Third, 
One limitation of this work is that the catchment attributes are fixed in time,
so we cannot yet detect changes within a basin such as a new dam, a wildfire, or
urban growth. %The agent marks these as needing external information. Fourth, the
%sliding windows overlap by four years, so rounds show a trend rather than
%independent samples, and NWM, whose record ends in 2023, covers only part of the
%most recent windows, which is exactly why HBV is used as the forward reference.
%The clearest next steps are a 
In addition to extending to other environmental domains, we will also consider  a forecasting setup with forecast inputs and the
addition of time-varying disturbance data.

\bibliographystyle{ACM-Reference-Format}
\bibliography{refs}

@article{addor2017camels,
  title={The CAMELS data set: catchment attributes and meteorology for large-sample studies},
  author={Addor, Nans and Newman, Andrew J and Mizukami, Naoki and Clark, Martyn P},
  journal={Hydrology and Earth System Sciences},
  volume={21}, number={10}, pages={5293--5313}, year={2017}
}

@inproceedings{chen2023physics,
  title={Physics-guided meta-learning method in baseflow prediction over large regions},
  author={Chen, Shengyu and Xie, Yiqun and Li, Xiang and Liang, Xu and Jia, Xiaowei},
  booktitle={Proceedings of the 2023 SIAM International Conference on Data Mining (SDM)},
  year={2023}
}

@inproceedings{yu2025survey,
  title={A survey of foundation models for environmental science},
  author={Yu, Runlong and Chen, Shengyu and Xie, Yiqun and Jia, Xiaowei},
  booktitle={Pacific-Asia Conference on Knowledge Discovery and Data Mining},
  pages={39--57},
  year={2025},
  organization={Springer}
}

@article{karpatne2018machine,
  title={Machine learning for the geosciences: Challenges and opportunities},
  author={Karpatne, Anuj and Ebert-Uphoff, Imme and Ravela, Sai and Babaie, Hassan Ali and Kumar, Vipin},
  journal={IEEE transactions on knowledge and data engineering},
  volume={31},
  number={8},
  pages={1544--1554},
  year={2018},
  publisher={IEEE}
}

@inproceedings{yu2025physics,
  title={Physics-guided foundation model for scientific discovery: An application to aquatic science},
  author={Yu, Runlong and Qiu, Chonghao and Ladwig, Robert and Hanson, Paul and Xie, Yiqun and Jia, Xiaowei},
  booktitle={Proceedings of the AAAI Conference on Artificial Intelligence},
  volume={39},
  number={27},
  pages={28548--28556},
  year={2025}
}

@inproceedings{chen2023meta,
  title={Meta-transfer-learning for time series data with extreme events: An application to water temperature prediction},
  author={Chen, Shengyu and Kalanat, Nasrin and Topp, Simon and Sadler, Jeffrey and Xie, Yiqun and Jiang, Zhe and Jia, Xiaowei},
  booktitle={Proceedings of the 32nd ACM International Conference on Information and Knowledge Management},
  pages={266--275},
  year={2023}
}

@article{nie2022time,
  title={A time series is worth 64 words: Long-term forecasting with transformers},
  author={Nie, Yuqi and Nguyen, Nam H and Sinthong, Phanwadee and Kalagnanam, Jayant},
  journal={arXiv preprint arXiv:2211.14730},
  year={2022}
}

@article{chen2023tsmixer,
  title={Tsmixer: An all-mlp architecture for time series forecasting},
  author={Chen, Si-An and Li, Chun-Liang and Yoder, Nate and Arik, Sercan O and Pfister, Tomas},
  journal={arXiv preprint arXiv:2303.06053},
  year={2023}
}

@inproceedings{cheng2026llm,
  title={Llm-based evaluation policy extraction for ecological modeling},
  author={Cheng, Qi and Liu, Licheng and Chen, Yixuan and Zhu, Qing and Yu, Runlong and Jin, Zhenong and Xie, Yiqun and Jia, Xiaowei},
  booktitle={Proceedings of the 32nd ACM SIGKDD Conference on Knowledge Discovery and Data Mining V. 2},
  pages={10704--10715},
  year={2026}
}

@inproceedings{chen2022physics,
  title={Physics-guided graph meta learning for predicting water temperature and streamflow in stream networks},
  author={Chen, Shengyu and Zwart, Jacob A and Jia, Xiaowei},
  booktitle={Proceedings of the 28th ACM SIGKDD conference on knowledge discovery and data mining},
  pages={2752--2761},
  year={2022}
}

@article{cheng2026hydrotft,
  title={HydroTFT: a cross-basin attention model for multi-horizon rainfall--runoff prediction},
  author={Cheng, Qi and Fan, Yingda and Jia, Xiaowei and Xie, Yiqun and Lu, Dan},
  journal={Machine Learning: Earth},
  volume={2},
  number={2},
  pages={025014},
  year={2026},
  publisher={IOP Publishing}
}

@article{fan2026zerodiff,
  title={ZeroDiff: Zero-Shot Time Series Reconstruction via Informed-Prior Diffusion},
  author={Fan, Yingda and Lu, Dan and Jia, Xiaowei},
  journal={arXiv preprint arXiv:2609.37078},
  year={2026}
}

@article{betancourt2021aq,
  title={AQ-Bench: a benchmark dataset for machine learning on global air quality metrics},
  author={Betancourt, Clara and Stomberg, Timo and Stadtler, Scarlet and Roscher, Ribana and Schultz, Martin G},
  journal={Earth System Science Data Discussions},
  volume={2021},
  pages={1--26},
  year={2021},
  publisher={G{\"o}ttingen, Germany}
}

@article{wang2026carbonglobe,
  title={CarbonGlobe: A Global-Scale, Multi-Decade Dataset and Benchmark for Carbon Forecasting in Forest Ecosystems},
  author={Wang, Zhihao and Ma, Lei and Hurtt, George and Jia, Xiaowei and Li, Yanhua and Li, Ruohan and Li, Zhili and Xu, Shuo and Xie, Yiqun},
  journal={Advances in Neural Information Processing Systems},
  volume={38},
  year={2026}
}

@article{cheng2026agroflux,
  title={AgroFlux: A Spatial-Temporal Benchmark for Carbon and Nitrogen Flux Prediction in Agricultural Ecosystems},
  author={Cheng, Qi and Liu, Licheng and Zhang, Yao and Hong, Mu and Xie, Yiqun and Jia, Xiaowei},
  journal={arXiv preprint arXiv:2602.01614},
  year={2026}
}

@inproceedings{sun2026x,
  title={X-methanewet: A cross-scale global wetland methane emission benchmark dataset for advancing science discovery with ai},
  author={Sun, Yiming and Chen, Shuo and Chen, Shengyu and Qiu, Chonghao and Liu, Licheng and Oh, Youmi and Malone, Sparkle L and McNicol, Gavin and Zhuang, Qianlai and Smith, Chris and others},
  booktitle={Proceedings of the 32nd ACM SIGKDD Conference on Knowledge Discovery and Data Mining V. 1},
  pages={2782--2793},
  year={2026}
}

@article{kratzert2019towards,
  title={Towards learning universal, regional, and local hydrological behaviors via machine learning applied to large-sample datasets},
  author={Kratzert, Frederik and Klotz, Daniel and Shalev, Guy and Klambauer, G{\"u}nter and Hochreiter, Sepp and Nearing, Grey},
  journal={Hydrology and Earth System Sciences},
  volume={23},
  number={12},
  pages={5089--5110},
  year={2019},
  publisher={Copernicus Publications G{\"o}ttingen, Germany}
}

@inproceedings{zeng2023transformers,
  title={Are transformers effective for time series forecasting?},
  author={Zeng, Ailing and Chen, Muxi and Zhang, Lei and Xu, Qiang},
  booktitle={Proceedings of the AAAI conference on artificial intelligence},
  volume={37},
  number={9},
  pages={11121--11128},
  year={2023}
}

@article{newman2015,
  title={Development of a large-sample watershed-scale hydrometeorological data set for the contiguous USA},
  author={Newman, Andrew J and Clark, Martyn P and Sampson, Kevin and others},
  journal={Hydrology and Earth System Sciences},
  volume={19}, number={1}, pages={209--223}, year={2015}
}

@article{kratzert2023caravan,
  title={Caravan--A global community dataset for large-sample hydrology},
  author={Kratzert, Frederik and Nearing, Grey and Addor, Nans and others},
  journal={Scientific Data}, volume={10}, number={1}, pages={61}, year={2023}
}

@article{tran2025camelsh,
  title={{CAMELSH}: A Large-Sample Hourly Hydrometeorological Dataset and Attributes at Watershed-Scale for {CONUS}},
  author={Tran, V. N. and Xu, D. and Nguyen, T. V. and Kim, T. and Ivanov, V. Y.},
  journal={Scientific Data}, volume={12}, pages={1307}, year={2025}
}

@article{kratzert2018,
  title={Rainfall--runoff modelling using Long Short-Term Memory ({LSTM}) networks},
  author={Kratzert, Frederik and Klotz, Daniel and Brenner, Claire and Schulz, Karsten and Herrnegger, Mathew},
  journal={Hydrology and Earth System Sciences},
  volume={22}, number={11}, pages={6005--6022}, year={2018}
}

@article{kratzert2019,
  title={Towards learning universal, regional, and local hydrological behaviors via machine learning applied to large-sample datasets},
  author={Kratzert, Frederik and Klotz, Daniel and Shalev, Guy and Klambauer, G{\"u}nter and Hochreiter, Sepp and Nearing, Grey},
  journal={Hydrology and Earth System Sciences},
  volume={23}, number={12}, pages={5089--5110}, year={2019}
}

@article{tft2023,
  title={Temporal Fusion Transformers for streamflow prediction: Value of combining attention with recurrence},
  author={Koya, Sadegh Sadeghi and Roy, Tirthankar},
  journal={Journal of Hydrology}, year={2024}, note={arXiv:2305.12335}
}

@article{nash1970,
  title={River flow forecasting through conceptual models part I---A discussion of principles},
  author={Nash, J. E. and Sutcliffe, J. V.},
  journal={Journal of Hydrology}, volume={10}, number={3}, pages={282--290}, year={1970}
}

@article{gupta2009,
  title={Decomposition of the mean squared error and {NSE} performance criteria: Implications for improving hydrological modelling},
  author={Gupta, Hoshin V and Kling, Harald and Yilmaz, Koray K and Martinez, Guillermo F},
  journal={Journal of Hydrology}, volume={377}, number={1-2}, pages={80--91}, year={2009}
}

@article{kling2012,
  title={Runoff conditions in the upper Danube basin under an ensemble of climate change scenarios},
  author={Kling, Harald and Fuchs, Martin and Paulin, Maria},
  journal={Journal of Hydrology}, volume={424}, pages={264--277}, year={2012}
}

@article{yilmaz2008,
  title={A process-based diagnostic approach to model evaluation: Application to the {NWS} distributed hydrologic model},
  author={Yilmaz, Koray K and Gupta, Hoshin V and Wagener, Thorsten},
  journal={Water Resources Research}, volume={44}, number={9}, year={2008}
}

@article{seibert2012hbv,
  title={Teaching hydrological modeling with a user-friendly catchment-runoff-model software package ({HBV light})},
  author={Seibert, Jan and Vis, Marc J P},
  journal={Hydrology and Earth System Sciences},
  volume={16}, number={9}, pages={3315--3325}, year={2012}
}

@article{mmbenchlive,
  title={{MMBench-Live}: A Continuously Evolving Benchmark for Multimodal Models},
  author={Anonymous}, journal={arXiv preprint arXiv:2607.01813}, year={2026}
}

@article{livehousets,
  title={{LiveHouse-TS}: An Open-world Living Benchmark for Time Series Foundation Models},
  author={Wen, Haomin and Zhou, Ziyu and Liu, Qingxiang and Zhong, Siru and Liang, Yuxuan},
  journal={arXiv preprint arXiv:2608.17299}, year={2026}
}

@article{arenabencher,
  title={{ArenaBencher}: Automatic Benchmark Evolution via Multi-Model Competitive Evaluation},
  author={Anonymous}, journal={arXiv preprint arXiv:2510.08569}, year={2025}
}

@article{deepfact,
  title={{DeepFact}: Co-Evolving Benchmarks and Agents for Deep Research Factuality},
  author={Anonymous}, journal={arXiv preprint arXiv:2603.05912}, year={2026}
}

@article{dynsurvey2025,
  title={Recent Advances in Large Language Model Benchmarks against Data Contamination: From Static to Dynamic Evaluation},
  author={Anonymous}, journal={arXiv preprint arXiv:2502.17521}, year={2025}
}

@article{oracleproto,
  title={{OracleProto}: Reproducible, Cutoff-Aligned Evaluation as an Alternative to Live Benchmarks},
  author={Anonymous}, journal={arXiv preprint arXiv:2605.03762}, year={2026}
}

@misc{nwm,
  title={The National Water Model},
  author={{NOAA Office of Water Prediction}},
  howpublished={\url{https://water.noaa.gov/about/nwm}}, year={2023}
}

@article{nearing2024global,
  title={Global prediction of extreme floods in ungauged watersheds},
  author={Nearing, Grey and Cohen, Deborah and Dube, Vusumuzi and Gauch, Martin and Gilon, Oren and Harrigan, Shaun and Hassidim, Avinatan and Klotz, Daniel and others},
  journal={Nature}, volume={627}, number={8004}, pages={559--563}, year={2024}
}

@article{caravanmultimet,
  title={Caravan MultiMet: Extending Caravan with Multiple Weather Nowcasts and Forecasts},
  author={Kratzert, Frederik and Gauch, Martin and Klotz, Daniel and Nearing, Grey},
  journal={arXiv preprint arXiv:2411.09459}, year={2024}
}

@article{frame2022extreme,
  title={Deep learning rainfall--runoff predictions of extreme events},
  author={Frame, Jonathan M and Kratzert, Frederik and Klotz, Daniel and Gauch, Martin and Shalev, Guy and Gilon, Oren and Qualls, Logan M and Gupta, Hoshin V and Nearing, Grey S},
  journal={Hydrology and Earth System Sciences}, volume={26}, number={13}, pages={3377--3392}, year={2022}
}

@article{ansari2024chronos,
  title={Chronos: Learning the Language of Time Series},
  author={Ansari, Abdul Fatir and Stella, Lorenzo and Turkmen, Caner and others},
  journal={Transactions on Machine Learning Research}, year={2024}
}

@inproceedings{das2024timesfm,
  title={A decoder-only foundation model for time-series forecasting},
  author={Das, Abhimanyu and Kong, Weihao and Sen, Rajat and Zhou, Yichen},
  booktitle={International Conference on Machine Learning (ICML)}, year={2024}
}

@article{swemera2025,
  title={{SWE-MERA}: A Dynamic Benchmark for Agentically Evaluating Large Language Models on Software Engineering Tasks},
  author={Anonymous}, journal={arXiv preprint arXiv:2507.11059}, year={2025}
}

@article{livekbench2026,
  title={Outrunning {LLM} Cutoffs: A Live Kernel Crash Resolution Benchmark for All},
  author={Anonymous}, journal={arXiv preprint arXiv:2602.02690}, year={2026}
}

@article{contamsurvey2025,
  title={A Survey on Data Contamination for Large Language Models},
  author={Anonymous}, journal={arXiv preprint arXiv:2502.14425}, year={2025}
}

@article{feng2020,
  title={Enhancing streamflow forecast and extracting insights using long-short term memory networks with data integration at continental scales},
  author={Feng, Dapeng and Fang, Kuai and Shen, Chaopeng},
  journal={Water Resources Research}, volume={56}, number={9}, pages={e2019WR026793}, year={2020}
}

\appendix
\section{Reproducibility and Implementation}
\label{app:repro}

\begin{table}[t]
\centering
\caption{EvoBench components are domain-general. The streamflow case study
instantiates each; the final column sketches analogues in other
environmental-modeling tasks.}
\label{tab:generalize}
\small
\begin{tabular}{p{0.21\linewidth}p{0.32\linewidth}p{0.34\linewidth}}
\toprule
Component & Streamflow (this paper) & Other domains \\
\midrule
Reference model & per-basin HBV & process model (e.g.\ TEM for CH$_4$,
chemical-transport for air quality) \\
Failure cells & basin attributes $\times$ flow period & site attributes $\times$
season/regime \\
Spatial gap & leave-cluster-out basins (PUB) & leave-region-out monitoring sites \\
Evolving loop & 5-year sliding water-year window & sliding window at the task's
cadence \\
Plausibility check & water balance ($Q/P\!\le\!1$) & mass/energy budget closure \\
\bottomrule
\end{tabular}
\end{table}

% \paragraph{Code.} The benchmark is a set of dependency-light Python modules
% (numpy/pandas; PyTorch for the learned models). The deterministic core
% (\texttt{metrics.py}, \texttt{hbv.py}, \texttt{protocols.py},
% \texttt{diagnostic\_slicer.py}, \texttt{grid2d.py}, \texttt{spatial\_cv.py},
% \texttt{rounds\_all.py}, \texttt{three\_conditions.py}, \texttt{loop3\_seeds.py})
% computes every score; the agentic layer
% (\texttt{adjudicate.py}, \texttt{dataqc.py}, \texttt{hypothesize.py}) only
% produces human-auditable labels. A cohort token selects the basin set
% (\texttt{gold}~=~162 at $\geq$90\% coverage; \texttt{0.8}~=~1{,}139 at $\geq$80\%)
% and tags all caches so runs at different scales do not collide.

\paragraph{Models.} The LSTM, EA-LSTM, and GRU use hidden size 128 and dropout
0.4, with a last-step linear readout. The Transformer is encoder-only with
$d_{\text{model}}{=}64$, 4 heads, 2 layers, feed-forward 128, and sinusoidal
positional encoding ($\sim$69k parameters). The TCN has three residual blocks of
dilated causal convolutions (64 channels, kernel 3, dilations 1, 2, 4, dropout
0.3). DLinear decomposes the window into trend (25-day moving average) and
seasonal parts, applies one linear map over time to each, and joins the result
with the static attributes in the output head. TSMixer uses two blocks of
alternating time-mixing and feature-mixing MLPs (dropout 0.2) followed by
temporal mean pooling. The simplified PatchTST splits the window into six
30-day patches, embeds each, and runs a 2-layer Transformer encoder
($d_{\text{model}}{=}64$, 4 heads, feed-forward 128, dropout 0.3) over the
patch tokens. All learned models take a 180-day sequence of five NLDAS-2 forcings
(\texttt{Rainf}, \texttt{Tair}, \texttt{SWdown}, \texttt{LWdown},
\texttt{PotEvap}) plus 18 static attributes and predict standardized specific
discharge (mm/day). Static attributes are concatenated to each timestep, except
that the EA-LSTM feeds them only to its input gate and DLinear adds them at the
output head. Training uses Adam ($10^{-3}$) with batch 256. % and a budget of oneepoch: the number of iterations is $\lceil N/256 \rceil$, where $N$ is the number of valid (basin, day) training samples, so each sample is seen once in expectation. This gives 5{,}279 iterations on the core fixed split and
% 32{,}936 on the extended one, 6{,}406--7{,}318 (core) and 40{,}737--47{,}097
% (extended) per sliding round, and one epoch over the training basins of each
% spatial fold. A single seed (0) is used unless stated otherwise. 
%The EA-LSTM recurrence is compiled with TorchScript for speed; this does not change the model. 
% Training runs on
% the Apple MPS backend; at inference, input windows are assembled on the CPU and
% the forward pass runs on MPS.

\paragraph{HBV reference.} A 10-parameter HBV-light model (snow, soil-moisture,
and two response reservoirs) is vectorized over a 60-sample Monte-Carlo ensemble
and the KGE$'$-best parameter set is cached per basin. It is calibrated on
the same data as the learned models, i.e., all data before the test window:
water years 1986--2010 for the fixed split and up to each round's cutoff in the
sliding rounds. The first calibration year is a warm-up excluded from the
objective, and every evaluation run starts with a one-year spin-up so that the
snow, soil, and groundwater stores are not cold-started on the first scored day.
Because it can be run on any evaluation window, HBV serves as the forward
reference in Eq.~\eqref{eq:skill}.

\paragraph{Interventions (closing the loop).} The interventions add two knobs,
each a response to a benchmark diagnosis: a \emph{low-flow-weighted loss}
($3\times$ weight on each basin's bottom-30\% flow days of the training window)
and $3\times$ \emph{over-sampling} of basins. In the fixed-split experiment the
over-sampled basins are the arid and snow basins, and each configuration is
trained with seeds 0--2 (one epoch each). In the three-condition experiment
the over-sampled basins are those in the previous round's promoted cells; the
baseline and targeted conditions train from scratch each round (one epoch),
while targeted+memory warm-starts from the previous round's model and trains
for 20\% of an epoch. There, input normalization is fixed from the round-1
training window so that a warm-started model sees consistent inputs. Only the
intervention knobs change between conditions; data, architecture, and seed are
identical.

\paragraph{Agentic layer.} Each agent follows the pattern ``deterministic core
emits an evidence bundle $\to$ LLM returns a strict-JSON verdict $\to$ a
deterministic rule proxy is the fallback.'' The data-quality auditor's decisive
checks (runoff ratio $>1 \Rightarrow$ exclude) are deterministic; only borderline
cases (e.g.\ long flatlines in regulated basins) are referred to the LLM judge.
Verdicts are stored as JSON and never enter any score.

\end{document}